\documentclass[conference]{IEEEtran}
\IEEEoverridecommandlockouts
\usepackage{cite}
\usepackage{amsmath,amssymb,amsfonts}
\usepackage{graphicx}
\usepackage[noend]{algpseudocode}
\usepackage[table]{xcolor}
\def\BibTeX{{\rm B\kern-.05em{\sc i\kern-.025em b}\kern-.08em
    T\kern-.1667em\lower.7ex\hbox{E}\kern-.125emX}}

\usepackage{algorithm}

\begin{document}

\title{Behavioral Repertoires for Soft Tensegrity Robots }

\author{\IEEEauthorblockN{Kyle Doney\IEEEauthorrefmark{2}, Aikaterini Petridou\IEEEauthorrefmark{2},  Jacob Karaul\IEEEauthorrefmark{2}, Ali Khan\IEEEauthorrefmark{2},  Geoffrey Liu\IEEEauthorrefmark{2} and   John Rieffel\IEEEauthorrefmark{1}}
\IEEEauthorblockA{
\IEEEauthorrefmark{1}
Computer Science Department\\
Union College\\
Schenectady, NY \\
Email: rieffelj@union.edu}
\IEEEauthorrefmark{2}
Undergraduate
}


\IEEEoverridecommandlockouts
\IEEEpubid{\makebox[\columnwidth]
{978-1-7281-2547-3/20/\$31.00~\copyright2020 IEEE \hfill} 
\hspace{\columnsep}\makebox[\columnwidth]{ }}
\IEEEpubidadjcol

\maketitle

\begin{abstract}

Mobile soft robots offer compelling applications in fields ranging from urban search and rescue to planetary exploration.  A critical challenge of soft robotic control is that the nonlinear dynamics imposed by soft materials often result in complex behaviors that are counter-intuitive and hard to model or predict.  As a consequence, most behaviors for mobile soft robots are discovered through empirical trial and error and hand-tuning.  A second challenge is that soft materials are difficult to simulate with high fidelity - leading to a significant {\em reality gap} when trying to discover or optimize new behaviors.  In this work we employ a Quality Diversity Algorithm running model-free on a physical soft tensegrity robot that autonomously generates a behavioral repertoire  with no {\em a priori} knowledge of the robot's dynamics, and minimal human intervention.   The resulting behavior repertoire displays a diversity of unique locomotive gaits useful for a variety of tasks.   These results help provide a road map for  increasing the behavioral capabilities of mobile soft robots through real-world automation. 

\end{abstract}

\section{Introduction}

One of the ``Grand Challenges'' of robotics is to ``translate fundamental biological principles into engineering design rules...to create robots that perform like natural systems''\cite{grandchallenges}.  In pursuit of that challenge, soft robotics seeks to use materials that more closely mimic the properties of natural systems
\cite{lipson2014challenges,wehner2016integrated,trimmer2008new,lin2011,shepherd2011multigait,martinez2013robotic}.      Unfortunately, soft-materials introduce considerable elasticity and deformability, resulting in robots with  nearly infinite degrees of freedom, and significant dynamical complexities.  As a result, it can often be difficult to find effective controllers for soft robots, particularly mobile soft robots\cite{lipson2014challenges,shepherd2011multigait,trimmer2008new}.   Consequently, most soft robotic locomotive behaviors are therefore developed by hand through empirical trial-and-error\cite{shepherd2011multigait}, and their actions tend to be limited to a single gait.  

This paper describes methods to {\em autonomously} discover novel and effective soft robotic locomotive behaviors without the biases and limitations of these human-in-the-loop approaches.  Moreover, we demonstrate that these methods can be run on a physical robot, without the need for simulation or modeling.  Specifically we show how Quality Diversity Algorithms (QDAs) can autonomously discover a diverse {\em repertoire} of unique locomotive behaviors for a tether-free soft tensegrity robot.  The discovered behaviors could, in turn, be incorporated into higher-level control strategies.    These methods offer to advance the state of the art in soft robotics by allowing researchers to more fully exploit the dynamical complexities of soft robots -- leading to more versatile and robust mobile soft robots. To the best of our knowledge, no other mobile soft robot has demonstrated as wide a variety of behaviors, much less autonomously generated behaviors.

\section{Background and Related Work}

The inherent complexities of soft materials makes soft robots incredibly difficult to simulate with sufficient fidelity to address the reality gap~\cite{jakobi1995noise}.  Several soft robotic simulators exist, including VoxCraft~\cite{liu_voxcraft_2020} and SOFA~\cite{duriez2013control}, both of which have produced transferable simulations only in very limited contexts~\cite{kriegman2020scalable,navarro2020model}.   The process of discovering locomotive behaviors for soft robots is often accomplished through empirical trial-and-error on physical robots\cite{shepherd2011multigait}, a method that can be both challenging and time consuming. As a consequence, the ensuing behaviors tend to either be slow, and quasi-static, or else largely uncontrolled \cite{trimmer2008new, lin2011,shepherd2011multigait,bartlett20153d}.   The use of  {\em ad hoc} and hard-coded controllers also means that these soft robots are generally unable to autonomously adapt to internal or external changes, for instance when they are physically damaged, or when they encounter a unknown terrain.   Moreover, many existing mobile soft robots tend to be designed only to perform one task, such as crawling or swimming, and exhibit little or no steering, turning, or any wide diversity of behaviors.

\subsection{Tensegrity-Based Soft Robots}

 \begin{figure}
      \centering
	\includegraphics[width=0.80\columnwidth]{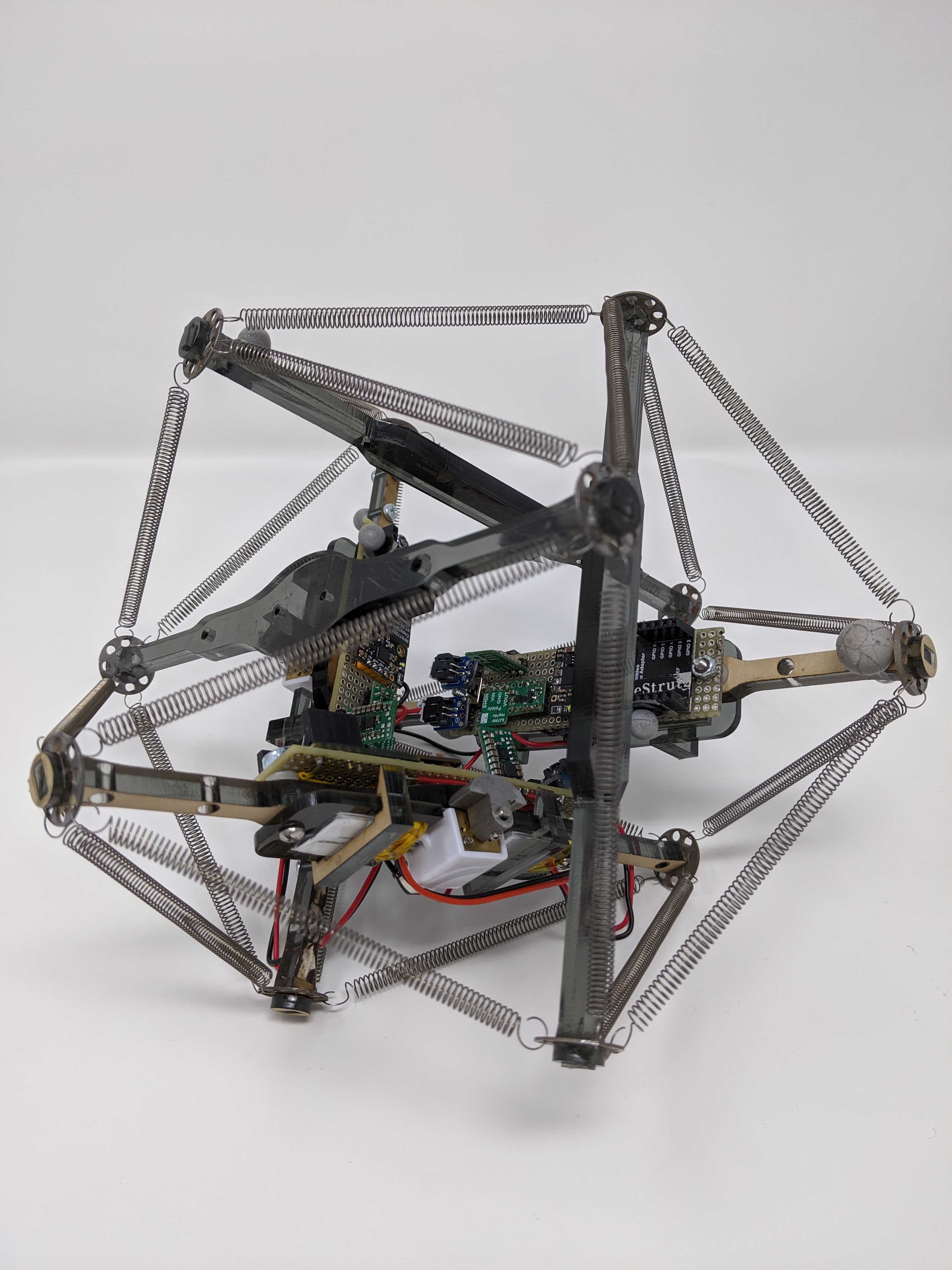}
	\caption{\label{fig:concept}
Tensegrity Robots exhibit many of the complex behaviors of more fully soft robots.  This robot, consists of 3 struts vibrational motors, 3 passive struts without motors, and 18 springs.}

\end{figure}

Tensegrities (Figure~\ref{fig:concept}) are relatively simple mechanical systems, consisting of a number of rigid elements (struts) joined at their endpoints by tensile elements (cables or springs), and kept stable through a synergistic interplay of pre-stress forces.  At every scale, tensegrity structures  have an impressive strength-to-weight ratio, are structurally robust, and stable in the face of deformation~\cite{snelson2012,skelton2009tensegrity}. Moreover, by choosing appropriate spring constants they can be used to construct {\em soft} robots     with a high dimensional configuration space~\cite{tibert2011review}, high  compliance\cite{bohm2017compliant} and  capable of rapid shape change and tunable stiffness ~\cite{tibert2002deployable}.   Unlike many other soft robots, tensegrity structures are inherently modular and are therefore, in principle, relatively easy to construct. 



As robots, tensegrity-based robots are typically controlled by changing the lengths of the tensile strings, causing large-scale deformations of the structure, which in turn cause the robot to step or tumble \cite{chandana-ieee,iscen2013controlling,iscen2014flop,agogino2013super,zappetti2017bio}, in a method NASA calls  `flop and roll,'.   More recently, researchers have begun to explore more dynamic methods of control.  Bliss {\em et al.} for instance \cite{bliss2012resonance,bliss2013central} use central pattern generators (CPGs) to control tensegrity-based swimming robots.  Compellingly, they were able to demonstrate a control technique based upon the resonant {\em entrainment} of the robot.   Mirletz {\em et al.} have used CPGs to produce goal-directed behavior in simulated tensegrity-spine-based  robots~\cite{mirletz2015goal}.     More recently, B{\"o}hm and Zimmermann developed a tensegrity-inspired robot actuated by an single oscillating electromagnet~\cite{bohm2013vibration}.

The appeal of using tensegrities as the basis for soft robotics research is that they are relatively inexpensive to build, they are inherently modular, and, like more fully soft robots they are capable of a  wealth  of dynamic behaviors including crawling, walking, tumbling and hopping.  The challenge of tensegrity robots, like soft robots, lies in finding ways to explore and exploit these behaviors.

In prior work, we describe the design of a relatively inexpensive and tether-free soft tensegrity robot 
\cite{kimber2019low} (Fig.~\ref{fig:concept}).  In contrast to many other soft robotic platforms, which require tethers for pneumatic or electrical power, this robot is capable of autonomous and tether-free operation -- allowing for a wide range of behaviors without the risk of tangling a tether.  The low cost, modularity, and ease of assembly help lower the barrier to entry for soft robotics research. 

Earlier work of ours also demonstrates how machine learning algorithms can be used to optimize gaits for the robot across level terrain~\cite{kimber2019low, khazanov2014evolution,khazanov2013exploiting}.  During this prior work  we observed that the robot is actually capable of a wide variety of translational and rotational behaviors and that the nonlinearity of the system means that the changes are consistent across trials and yet non-intuitive.  The diversity of these behavioral modalities far exceeds those typically seen in soft robots.  It is this last insight that guides our inquiry: how best to explore and exploit the entire range of a soft robot's behavioral diversity.

\subsection{Quality Diversity Algorithms}

\begin{figure*}[t!]
\begin{center}
\includegraphics[height=0.30\linewidth]{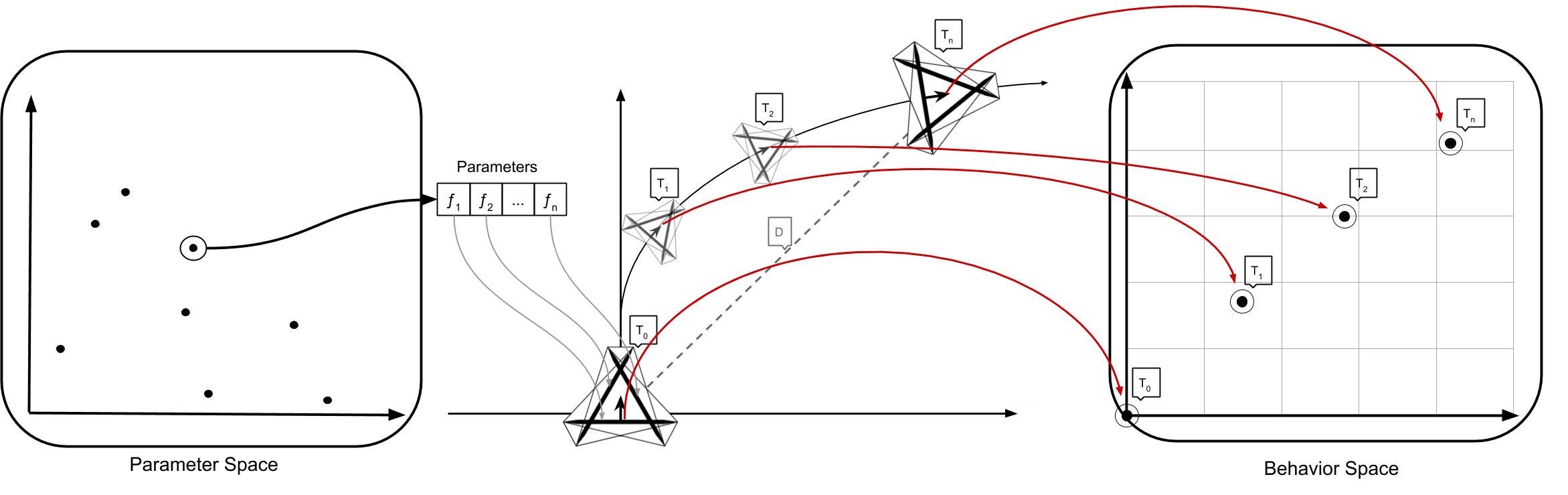}
\end{center}
\caption{Quality Diversity algorithms generate a mapping between parameter space $\mathcal{P}$ and behavior space $\mathbb{B}$. Typically, the behavior space is discretized into a set of bins, and when necessary a {\em quality metric} is used to compare behaviors in said bins.}\label{fig:dynamic-qd}
\end{figure*}

Quality Diversity algorithms (QDAs)~\cite{stanley-qd,duarte2018evolution,lehman2011evolving,cully2015robots} search for {\em novelty} alongside {\em optimality} in a given space.   When applied to robotics, this can mean generating mappings between a robot's {\em parameter space} (those features that can be directly affected by a controller), and its {\em behavior space} (a description of the outcomes of those actions), as illustrated by Figure~\ref{fig:dynamic-qd}.  The outcome of a successful quality diversity search is a {\em behavioral repertoire} describing a diverse variety of reachable behaviors that the robot is capable of -- essentially a mapping between parameter space and behavior space.  QDAs have been employed with considerable success with both simulated virtual agents~\cite{lehman2011evolving, duarte2018evolution} and physical robots~\cite{cully2015robots}.  Most notably, Cully~{\em et al.}~\cite{cully2015robots} used a hybrid simulation/physical approach to behavior build repertoires for a hexapod robot.  

However, most soft robots, including our tensegrity robots, are in principle capable of a considerably wider range of behaviors than those exhibited by the systems in the above work, due to their soft material composition and wide range of degrees of freedom.  Moreover, the complexity of soft systems, including tensegrities, means that repertoires cannot be developed in simulation and then transferred into physical system with any degree of expected fidelity~\cite{jakobi1995noise}.  In this work, we show how quality diversity algorithms can be effectively leveraged to generate behavioral repertoires for soft robotics operating in the real world.

\section{Behavioral Repertoires for Tensegrity Robots}

In the context of our tensegrity robots, we use the MAP-Elites quality diversity algorithm to generate a behavioral repertoire for the tensegrity robot ~\cite{cully2015robots}.  In particular, the tensegrity robot has three vibrational motors each of which can be assigned a frequency, and so a set of parameters $p \in \mathcal{P}$ (referred to as Map of parameters $\mathcal{P}$  in Algorithm~\ref{alg:mapelites}) can be expressed as $p = (f_1, f_2, f_3)$ where $f_i \in \mathbb{Z}$ (integers) and $255 \geq f_i \geq 0$. A locomotive behavior of the robot is described by the robot's displacement in the Cartesian (x/y) coordinate plane and its rotation around the z axis (yaw) over an evaluation period. Formally a behavior $b$ in the behavior space $\mathbb{B}$ to equal $(\Delta X, \Delta Y, \Delta \Psi)$ where $\Delta X$ and $\Delta Y$ are within the range $-360$mm:$360$mm, and $\Delta \Psi$ is within the range $-180^\circ$:$180^\circ$. Reasonable bounds on the limits of the behavior space were determined empirically based on the displacements achieved in prior experimentation and are equivalent to 3.5 body-lengths of our robot.

As an example of this behavior parameterization, a set of motor parameters $p = (189, 30, 251)$ that caused the robot to displace 50mm along the x axis and 25mm along the y axis, without rotating, would result in $b = (50,25,0)$.   And a robot that spun $90^{\circ}$ without displacement would result in $b = (0,0,90^{\circ})$. 

When discretizing the behavior space into intervals (bins), we address the stochasticity of our robot by choosing a interval size based on standard deviation testing. When considering the usefulness of a behavior repertoire, it is important to maximize the probability that if set $p$ generated behavior $b$ in bin $x$, then any further evaluating of $p$ also generate $b^\prime \in x$. This repeatability testing consisted of re-running the same set $p \in \mathcal{P}$ on the robot. For each $p$, we ran 30 tests: 10 five second tests, 10 ten second tests, and 10 fifteen second tests. We tested 15 parameter sets that represented a diverse selection of the behavior space.
Based on the results of this testing, we discretized $\mathbb{B}$ into 864 bins. The x displacement and y displacement axes of $\mathbb{B}$ were discretized into 12 bins at 6cm intervals, and the $\psi$ displacement axis $\mathbb{B}$ was discretized into 6 bins at 60 degree intervals. We also determined that 10 second trials were optimal to maximize repeatability of behaviors, while also minimizing trial length.

\begin{algorithm} [h!]
    \caption{MAP-Elites, Modified from~\cite{mouret_illuminating_2015}}
    \label{alg:mapelites}
    \begin{algorithmic}
      \Function{MAP-Elites}{$fitness()$, $variation()$, $\mathcal{X}_{initial}$}
      \State $\mathcal{P}\gets\emptyset$, $\mathcal{F}\gets\emptyset$
      \Comment{\textit{Map of params $\mathcal{P}$, and fitnesses $\mathcal{F}$}}

      \State $\mathit{B_{initial}} \gets behavior\_descriptor(\mathcal{P}_{initial})$
      \State $\mathcal{P}(B_{initial}) \gets \mathcal{P}_{initial}$
      \Comment{\textit{Place initial solutions in map}}
      \State $\mathcal{F}(B_{initial}) \gets fitness(B_{initial}))$

        \For{iter = $1 \to I$}
          \State $\mathbf{p'} \gets variation(\mathcal{P})$ \Comment{\textit{Create new solns from elites}}
          \State $\mathbf{b'} \gets behavior\_descriptor(\mathbf{p'})$
          \State $\mathbf{f'} \gets fitness(\mathbf{p'})$
          \\ \Comment{\textit{Replace if better}}
          \If{$\mathcal{F}(\mathbf{b'}) = \emptyset$ or $\mathcal{F}(\mathbf{b'}) < \mathbf{f'}$}
            \State $\mathcal{F}(\mathbf{b'}) \gets \mathbf{f'}$
            \State $\mathcal{P}(\mathbf{b'}) \gets \mathbf{p'}$
          \EndIf
        \EndFor
        \State \Return $(\mathcal{P}$, $\mathcal{F})$
        \Comment{\textit{Return illuminated map}}
        \EndFunction
    \end{algorithmic}
  \end{algorithm}
  

At a high-level the MAP-Elites generates the behavior repertoire $\mathcal{P}$ in two phases. The first phase consists of an initial random sampling of the parameter space, where sets of parameters $p \in \mathcal{P}$ are randomly selected and evaluated on the robot. The resulting behaviors (the elites) are stored as mappings in the behavior repertoire $\mathcal{P}$. The second phase of the algorithm selects an elite previously stored$\in \mathcal{P}$, slightly mutates the parameter set associated with the elite, then evaluates the new parameter set on the robot. 

If in either phase a set of parameters is evaluated where multiple behaviors occupy the same bin, a fitness function (quality metric) is employed to select the "better" behavior. Our {\em fitness metric} $f(b) = |\Delta x| + |\Delta y| + |\Delta \phi|$ selects competing behaviors within a niche with higher absolute displacement, indicating a preference for behaviors at the more extreme end of each behavioral niche. In other words, $f$ optimizes each behavior bin $\in \mathbb{B}$ so that the most active behaviors of the robot are not discarded. All three components of $b$ have be normalized to the range 0-1, so they factor equally into the {\em fitness metric}.

As opposed to a method such as random selection, by employing this fitness function the algorithm ensures that the more time-efficient behaviors are kept. In other words, behaviors which displace the robot more over the same time-period are inherently more interesting from a use-ability standpoint. While the primary goal of the QDA in this instance is to build a repertoire of diverse behaviors (fill many unique bins on the map), the quality of each kept behavior is still important. During automatic behavior generation it is much preferred that each of locomotive behaviors in the constructed behavior set is the optimal behavior (the one with the greatest displacement) in that grid section, as this behavior maximizes robot displacement over time.

\begin{figure}[h]
    \centering
    \includegraphics[width=\columnwidth]{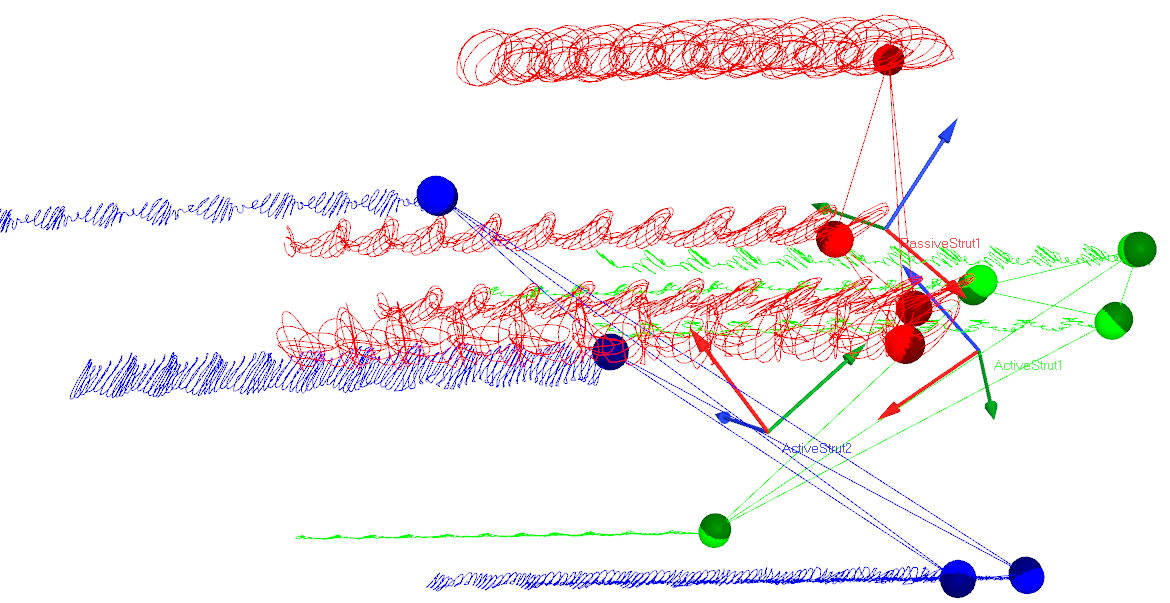}
    \caption{A 3D rigid body model close-up view of the dynamics of the tensegrity robot's locomotion. Each dot represents a marker being tracked with sub-millimeter accuracy on the tensegrity robot.  The resonances imposed by the three vibrating motors can be clearly seen.}
    \label{fig:zoomed}
\end{figure}

The displacement of the robot was tracked using a 20-camera Qualisys Oqus 700+ system and the QTM Tracking Software at a frame rate of 300 frames per second.   Four tracking markers placed on each of three of the tensegrity struts to provide 6DOF position and rotation data for each of the marked struts. This tracking setup can been seen in Figure \ref{fig:zoomed}.  Global reference frame tracking data was then converted into the robot's local reference frame in order to produce a behavior descriptor as described above. The "front" of the robot, corresponding to a global reference frame yaw of $0$ is essentially arbitrary (because the robot is rotationally symmetric), and so was determined by using the reported Euler angle of one of the tracked struts. 

\section{Experiment Design}

This experiment seeks to quantitatively determine if the QDA MAP-Elites is a data-efficient method for illuminating the behavior space of a tensegrity robot. This is done by comparing MAP-Elites illumination against a pseudo-random control.  A control computer running the MAP-Elites algorithm was used to generate frequencies that were sent to the robot's struts' Arduino controllers via Bluetooth.  Details of the construction of the robot, and its wireless control are provided in\cite{kimber2019low}.  An initial 100 random parameter trials were run (MAP-Elites phase 1). These trials were shared between both the control experiment and MAP-Elites experiment to ensure an equal baseline. If this was not done, the randomness of these initial seed trials could more greatly sway how well MAP-Elites performs in comparison with the control. Each time a unique behavior is found, it becomes slightly more difficult to find the next novel behavior, so if one experiment continued with 25 discovered behaviors, where the other continued with 60 behaviors, this would heavily bias the results.  

\begin{figure}[ht]
\centering
\includegraphics[width=1\linewidth]{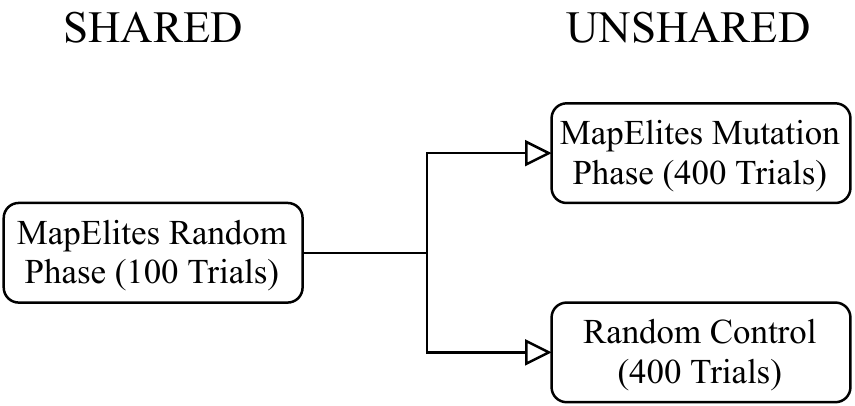}\\
\caption{This is a visual flowchart of the MAP-Elites vs. Control experiment. An initial amount of random trials are shared as to generate the initial behavior repertoire for MAP-Elites. This allows the mutation phase of the MAP-Elites algorithm to be directly compared with the random search.}\label{fig:Trial_Breakdown}
\end{figure}

The behavior repertoire generated from the shared 100 trials is used as the initial behavior repertoire for the second half of the experiment. The random control evaluates another 400 pseudo-randomly generated parameters on the robot. The non-control experiment continues by running the mutation phase of MAP-Elites (phase 2). This also evaluates another 400 sets of parameters on the tensegrity robot. A visual depiction of this is displayed in Figure \ref{fig:Trial_Breakdown}. 

Overall, the whole experiment required 900 robot trials, each of which consisted of 10 seconds of motor run-time, and an additional amount of cooldown time. This equates to approximately 3.5 hours of operating time. It is important to note that the number of trials for both sections of this experiment was empirically chosen. From some prior testing, it appears that increasing the number of trials during the initial random phase does not significantly change the outcome of the experiment. More work is needed to determine a lower-bound for the number of required initial random trials. Such a lower-bound number could be based on the percentage of bin's occupied in the behavior space.

\section{Results}

At the conclusion of our behavioral repertoire exploration our algorithm discovered 110 distinct behaviors during the non-control section of the experiment. This represents approximately $13\%$ of the discritized behavior space. The same amount of control trials discovered only 71 distinct behaviors or $8\%$ of the behavior space. To fully understand to what extent the MAP-Elites algorithm out performed the random control, we must compare specifically the unshared sections of the experiment. In these trials MAP-Elites discovered 83 unique behaviors, compared to the control discovering 44. In other words, the mutation phase of MAP-Elites found unique behaviors twice as fast as the control. The average quality of the distinct behaviors found by MAP-Elites is statistically significantly higher than the average quality of control behaviors. What this means is that the behaviors generated by the MAP-Elites mutation have on average larger displacement values. Table I summarizes the numeric data generated by this experiment. 

\begin{table*}[ht]
\centering
\begin{tabular}{|l|l|l|l|l|}
\hline
\rowcolor[HTML]{C0C0C0} 
{\color[HTML]{333333} Trial Sets:} &
  {\color[HTML]{333333} \begin{tabular}[c]{@{}l@{}}Initial 100 Random \\ (Phase 1)\end{tabular}} &
  {\color[HTML]{333333} \begin{tabular}[c]{@{}l@{}}Mutation 400 \\ (Phase 2)\end{tabular}} &
  {\color[HTML]{333333} \begin{tabular}[c]{@{}l@{}}Random 400 \\ (Phase 2)\end{tabular}} &
  {\color[HTML]{333333} \begin{tabular}[c]{@{}l@{}}Mutation 500\\  (Phase 1 + 2)\end{tabular}} \\ \hline
\cellcolor[HTML]{C0C0C0}Number of Unique Behaviors          & 27  & 83  & 44  & 110  \\ \hline
\cellcolor[HTML]{C0C0C0}Avg. Fitness of Unique Behaviors& .67 & .97 & .87 & 0.90 \\ \hline
\cellcolor[HTML]{C0C0C0}Avg. Fitness of all Trials       & .32 & .75 & .36 & 0.66 \\ \hline
\end{tabular}\label{tab:results}
\caption{A subset of the data collected from the MAP-Elites vs. Random search experiment. It includes the number of unique behaviors (number of bins occupied in the behavior space) found during each section of the experiment, and the average quality of those trials, based on the fitness metric.}
\end{table*}

\begin{figure*}[ht]
\includegraphics[width=1.00\linewidth]{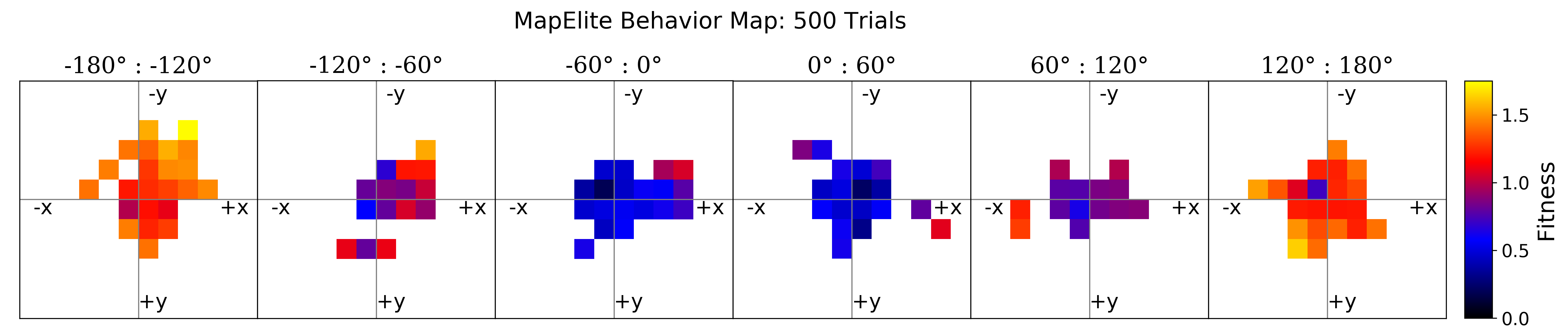}\\
\caption{A behavior map showing the diversity of behaviors discovered by the non-control MAP-Elites algorithm. Each graph displays the $\Delta x$ and $\Delta y$ displacements for all the behaviors within a specific rotational $\phi$ bin of the behavior map.  The furthest left graph, for instance, illustrates all the behaviors discovered for which the robot rotated between  $-120^{\circ}$ and $-180^{\circ}$ (negative rotations are anti-clockwise from initial position), and the adjacent graph illustrates behaviors for which the robot rotated between  $-60^{\circ}$ and  $-120^{\circ}$. The color of a square maps to the fitness value of the behavior in that bin.
}\label{fig:rotation-histos}
\end{figure*}

\begin{figure}[ht]
\centering
\includegraphics[width=0.8\columnwidth]{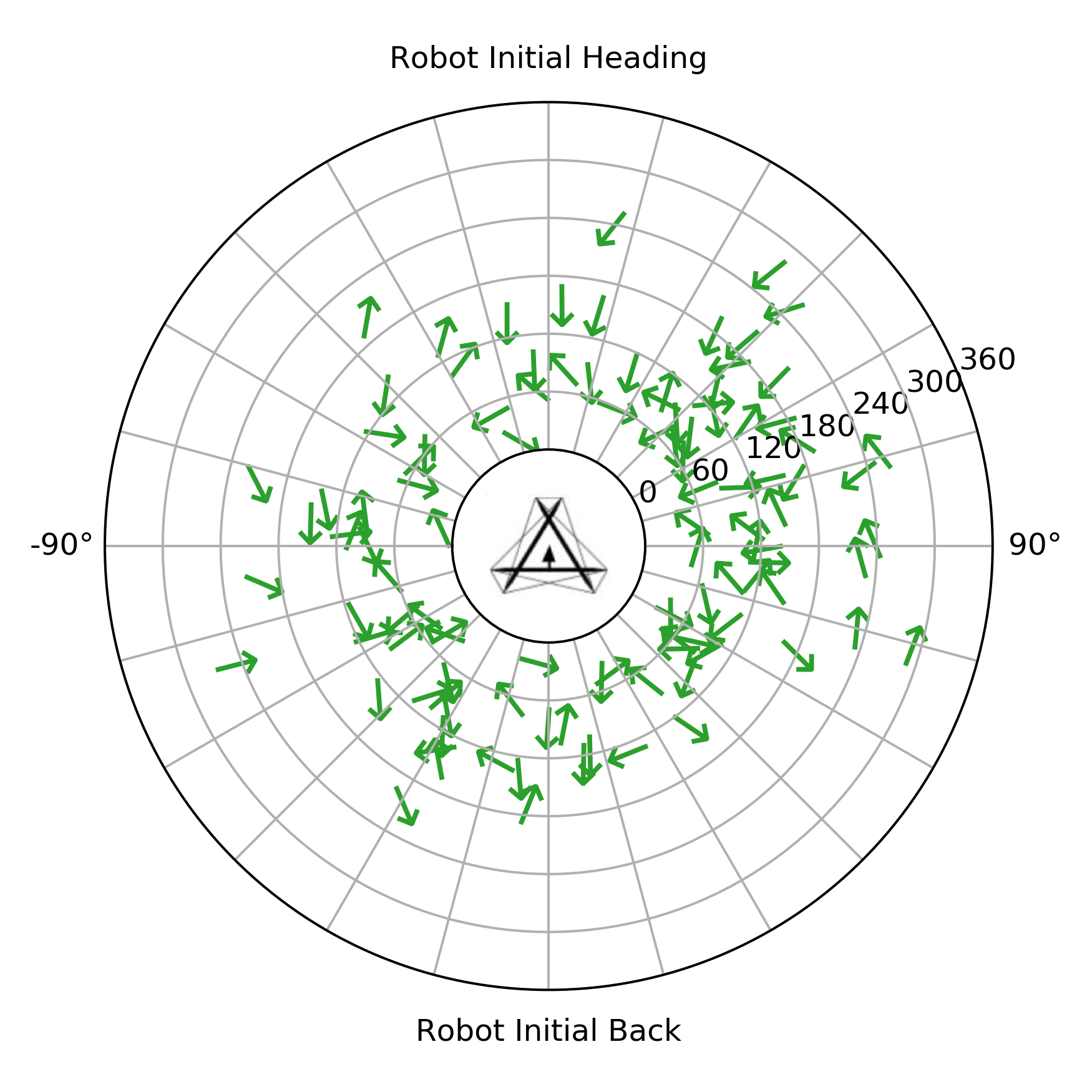}
\caption{
A top-down view of the behavior spaced overlaid in the robot's coordinate frame. Each green arrow represents the location and orientation of the robot after a behavior $b \in \mathbb{B}$ is run for 10 seconds. This figure shows the behaviors discovered by the non-control MAP-Elites algorithm.
}\label{fig:topdown}
\end{figure}

Figure \ref{fig:rotation-histos} illustrates the diversity of the behaviors produced. Not surprisingly, many of the behaviors discovered within the repertoire produce minimal motion: motor frequency parameter sets $p$ can contain very low motor speeds that produce little or no displacement or rotation of the robot. This is somewhat mitigated by only running the experiment if a threshold value is reached. If the three motor speeds added together do not exceed the threshold value, then it is assumed the robot will remain stationary. Performing more complex thresholding to remove more parameters which produce minimal motion would be exceedingly difficult due to the counter-intuitive nature of tensegrity dynamics.

The most qualitatively interesting behaviors are less common:  those behaviors that result in large $\Delta x \Delta y$ displacements are found somewhat more often at the outer edges of the high (positive or negative) rotation angle graphs (those being the leftmost and rightmost graphs of Figure~\ref{fig:rotation-histos}). It might be expected that those behaviors that translate the robot the furthest are less likely to rotate the robot, and vice versa, but this is not the case: many of the behaviors that produce a large translation also produce a correspondingly large rotation.

Another way to visualize this data is displayed in Figure~\ref{fig:topdown}. The tail of each arrow indicates the displacement of a specific behavior within its coordinate frame, and the orientation of the arrow indicates the behavior's rotation $\Delta \phi$.  The upper right most arrow, for instance, illustrates a behavior that caused the robot to displace forwards just around 240mm, and rotate approximately $+40^{\circ}$ (clockwise). What this visualization demonstrates is that distinct behaviors were discovered by our algorithm which move the robot in all 360 degrees. This demonstrates how behavior repertoire can be very easy converted into a robotic controller, as we already have known parameter sets which move the robot in essentially all directions.

\begin{figure*}
\centering
\includegraphics[height=0.13\linewidth]{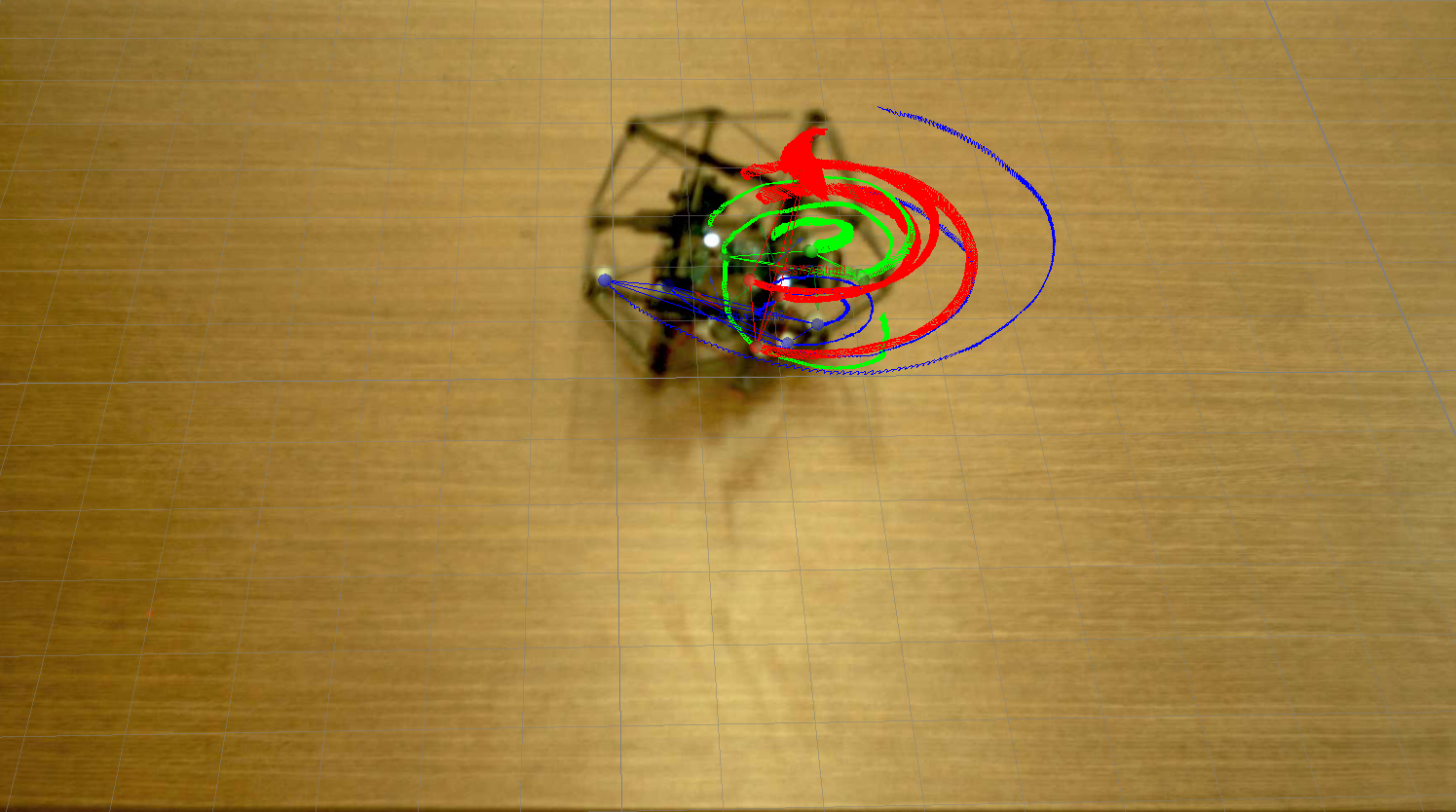} \hfill
\includegraphics[height=0.13\linewidth]{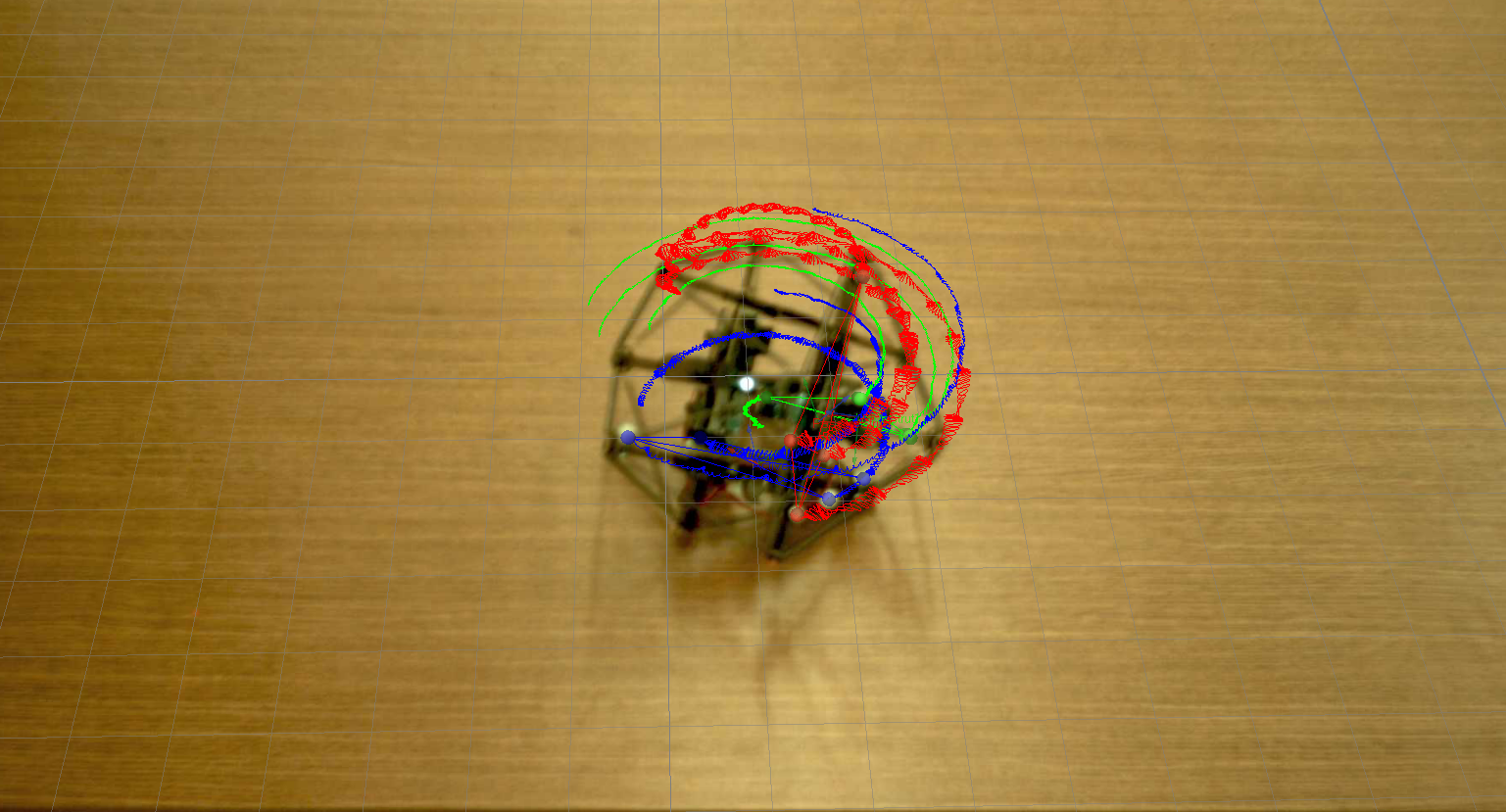} \hfill
\includegraphics[height=0.13\linewidth]{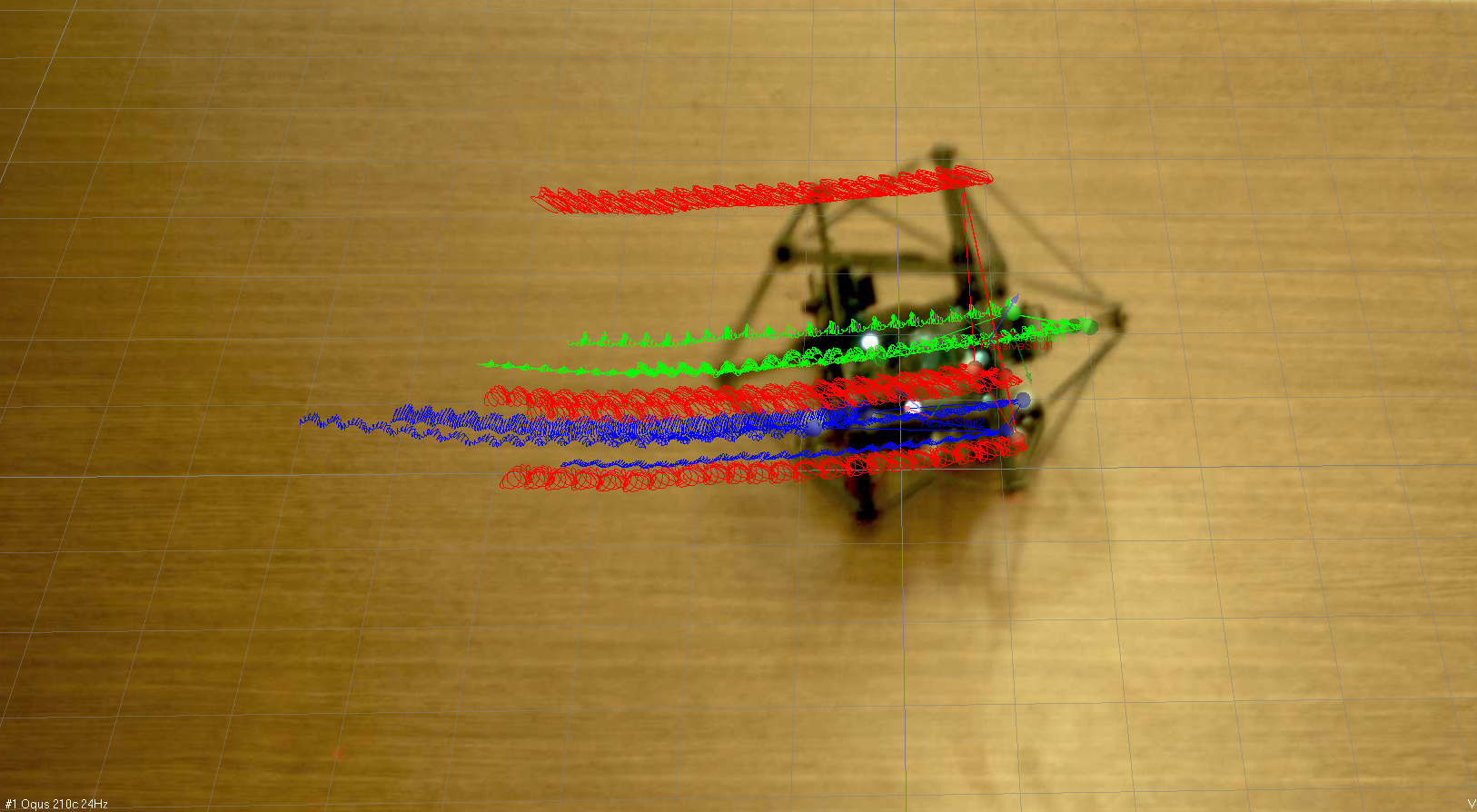} \hfill
\includegraphics[height=0.13\linewidth]{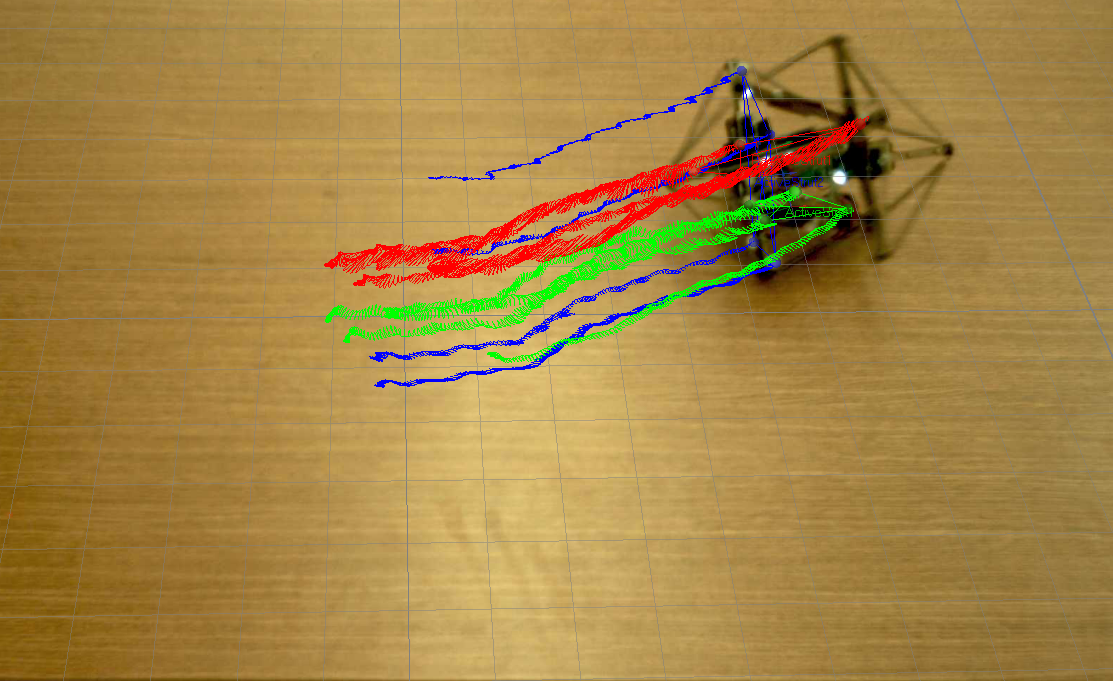} 
\caption{Overhead camera views and rigid body model overlays of several of the diverse behaviors discovered by our QDA.   In the two left-hand figures, the robot rotates with minimal displacement in the x/y plane.   In the two right-hand images, the robot translates with relatively little rotation. }\label{fig:spinny}
\end{figure*}

Figure~\ref{fig:spinny} provides examples of some of the behaviors discovered during our search.  Each behavior is both quantitatively and qualitatively distinct. The colored tracks on each image show the rigid model overlay for the three tracked struts.  The four markers per strut all share a common color. The two behaviors on the left hand side are both high rotation behaviors, with relatively low translation on the plane.  The behaviors on the right, by contrast, produce linear translation (although along different vectors), with minimal rotation.  

It is worth leveraging the 300fps frame rate and sub-millimeter accuracy of the Qualisys motion capture system in order to scrutinize the details of the robot's motion in high resolution.  Figure~\ref{fig:zoomed} provides this. The three motors in this gait are spinning at different frequencies, and we can see the periodic trace of the how those tracked points on the tensegrity robot vibrate.

\section{Discussion}

Our ambition in this paper is to demonstrate that Quality Diversity Algorithms (QDAs) running on real-world soft robots can, with little {\em a priori} knowledge, autonomously illuminate and exploit the tremendous diversity exhibited by soft robots.  The range and variety of behaviors discovered by these methods are significantly more dynamic and diverse than those demonstrated by other mobile soft robots.

Beyond the dynamics and algorithm, a design detail of our robot that facilities this diversity is the fact that unlike almost all other soft robots it is wireless and completely tether-free, unlike many mobile soft robots that are pneumatically actuated and tethered to externap pressure and electrical sources, and thereby limited in their ability to roll, twist, and turn in interesting ways.  Because our robot is entirely tether free and controlled via Bluetooth, we are able to explore a wider range of behaviors (indeed, earlier tethered versions of our robot became quite tangled in their tethers when rotating~\cite{khazanov2014evolution} ).

An added value of this process is that the automated nature of our algorithm is not confined by human intuition when exploring the behavior space, and is therefore significantly less fettered by human biases - for instance a preference for "forwards" behaviors (indeed our robot has no pre-defined front).   Small changes to the motor speeds of our robot can produce profound and unexpected changes in behavior -- for instance a switch from forward locomotion mode to a purely rotational mode.  Recent work of ours optimized a single behavior such as forward locomotion~\cite{rieffel2017soft, kimber2019low}, however, these methods by nature discard interesting behaviors that are non-optimal in regards to a single fitness function but that might otherwise be useful in some other context -- for instance rotation, or a behavior that mixes rotation and translation.  Indeed, in our earlier approaches to optimizing locomotion we discarded any behavior that produced any gaits that happened to also rotate the robot outside of a narrow window of angles.   This QDA approach, by contrast, allows for a balance between a diversity of behaviors (across different discretized bins of the behavior space) and optimization of behaviors (within bins). 


This raises the question of how exactly to make use of this repertoire.  Fortunately recent work on QDAs on more conventional robots in both simulation and the real world show how behavioral repertoires can be incorporated into high-level goal oriented tasks through methods such as policy networks~\cite{duarte2018evolution} or Monte Carlo Tree Search~\cite{chatzilygeroudis2018reset}.  Our interest going forward, therefore, is to work toward applying these hybrid QDA-planning methods toward the signicantly more complex behavior space of soft robots.

\section{Conclusions}


In this work we have illustrated how Quality Diversity Algorithms (QDAs) can be used to replace hand-tuning with automated locomotive behavior discovery.  Moreover, we demonstrated their utility on physical robots, without the need for low-fidelity soft robotic simulators and with minimal human intervention.   The QDA can relatively efficiently explore the large and complex search space of possible locomotive behaviors for the given soft robot.  The behavioral repertoire developed through this process shows a diversity rarely seen in other mobile soft robots, which tend to focus on one particular task domain such as forward locomotion.   The long-term outcome of this thread of inquiry will lead to new methods by which increasingly complex soft robots can not only outperform conventional robots across a range of environments, but also learn and resiliently adapt to changing conditions and damage. 

Many soft organisms in nature quite effectively leverage their material properties:  {\em Manduca sexta} caterpillars, for instance, have a mid-gut which acts like a ``visceral-locomotory piston''  to assist in locomotion~\cite{simon2010visceral}.  Jellyfish use their elasticity in order to recover energy during swimming \cite{gemmell2013passive}.   Understanding how to exploit soft material properties, and developing tools to explore their dynamics are both critical tasks for advancing the state of the art.
Our results are therefore an important step toward the soft robotic grand challenges of developing engineered systems that are are capable of the type of behavioral complexity seen in the natural world.  


\section*{ackowledgements}
This material is based upon work supported by the National Science Foundation under Grant Nos. NSF-1939930 and NSF-1827495.

\bibliographystyle{IEEEtran}
\bibliography{tensegrity}

\addtolength{\textheight}{-12cm}   




\end{document}